\documentclass[pdflatex, sn-mathphys]{sn-jnl}

\usepackage[english]{babel}
\usepackage{subfig}

\usepackage{siunitx}

\usepackage{titlesec}
\titlespacing*{\section}{0pt}{1.75ex plus .5ex minus .5ex}{-0.25ex plus .5ex minus 1ex}
\titlespacing*{\subsection}{0pt}{1.25ex plus .5ex minus .5ex}{-0.25ex plus .5ex minus 1ex}
\titlespacing*{\subsubsection}{0pt}{1ex plus .5ex minus .5ex}{0ex plus .25ex minus .25ex}
\titlespacing*{\paragraph}{0pt}{.75ex plus .25ex minus .25ex}{0ex plus .25ex minus .25ex}
\titlespacing*{\subparagraph}{0pt}{.5ex plus .25ex minus .25ex}{0ex plus .25ex minus .25ex}

\usepackage{array}

\newcolumntype{C}[1]{>{\centering\let\newline\\\arraybackslash\hspace{0pt}}m{#1}}
\newcolumntype{L}[1]{>{\raggedright\let\newline\\\arraybackslash\hspace{0pt}}m{#1}}
\newcolumntype{R}[1]{>{\raggedleft\let\newline\\\arraybackslash\hspace{0pt}}m{#1}}
\usepackage{etoolbox}
\makeatletter
\patchcmd{\ps@headings}
{\hbox to \hsize{\hfill Preprint sent to Applied Intelligence\hfill}}
{\hbox to \hsize{\hfill Preprint sent to Applied Intelligence\hfill}}
{}
{}
\patchcmd{\ps@titlepage}
{\hbox to \hsize{\hfill Preprint sent to Applied Intelligence\hfill}}
{\hbox to \hsize{\hfill Preprint sent to Applied Intelligence\hfill}}
{}
{}
\makeatother

\jyear{2022}%

\begin{document}

\title[Learning more with the same effort]{Learning more with the same effort: how randomization improves the robustness of a robotic deep reinforcement learning agent}

\author*[1]{\fnm{Luc\'{i}a} \sur{G\"{u}itta-L\'{o}pez}}\email{lucia.guitta@iit.comillas.edu}

\author[1]{\fnm{Jaime} \sur{Boal}}\email{jaime.boal@iit.comillas.edu}

\author[1]{\fnm{\'{A}lvaro J.} \sur{L\'{o}pez-L\'{o}pez}}\email{alvaro.lopez@iit.comillas.edu}

\affil[1]{\orgdiv{Institute for Research in Technology (IIT)}, \orgname{ICAI School of Engineering, Comillas Pontifical University}, \orgaddress{\street{Santa Cruz de Marcenado - 26}, \city{Madrid}, \postcode{28015}, \state{Madrid}, \country{Spain}}}


\abstract{The industrial application of Deep Reinforcement Learning (DRL) is frequently slowed down because of the inability to generate the experience required to train the models. Collecting data often involves considerable time and economic effort that is unaffordable in most cases. Fortunately, devices like robots can be trained with synthetic experience thanks to virtual environments. With this approach, the sample efficiency problems of artificial agents are mitigated, but another issue arises: the need for efficiently transferring the synthetic experience into the real world (sim-to-real).

This paper analyzes the robustness of a state-of-the-art sim-to-real technique known as progressive neural networks (PNNs) and studies how adding diversity to the synthetic experience can complement it. To better understand the drivers that lead to a lack of robustness, the robotic agent is still tested in a virtual environment to ensure total control on the divergence between the simulated and real models. 

The results show that a PNN-like agent exhibits a substantial decrease in its robustness at the beginning of the real training phase. Randomizing certain variables during simulation-based training significantly mitigates this issue. On average, the increase in the model's accuracy is around 25\% when diversity is introduced in the training process. This improvement can be translated into a decrease in the required real experience for the same final robustness performance. Notwithstanding, adding real experience to agents should still be beneficial regardless of the quality of the virtual experience fed into the agent.

This article was accepted and published in Applied Intelligence. Cite as: G\"{u}itta-L\'{o}pez, L., Boal, J. \& L\'{o}pez-L\'{o}pez, \'{A}.J. Learning more with the same effort: how randomization improves the robustness of a robotic deep reinforcement learning agent. Appl Intell 53, 14903–14917 (2023). https://doi.org/10.1007/s10489-022-04227-3}




\maketitle

\section{Introduction}\label{sec1}
\begin{figure}[h]
	\centering
	\renewcommand{\figurename}{Fig.}
	\includegraphics[width=0.3\textwidth]{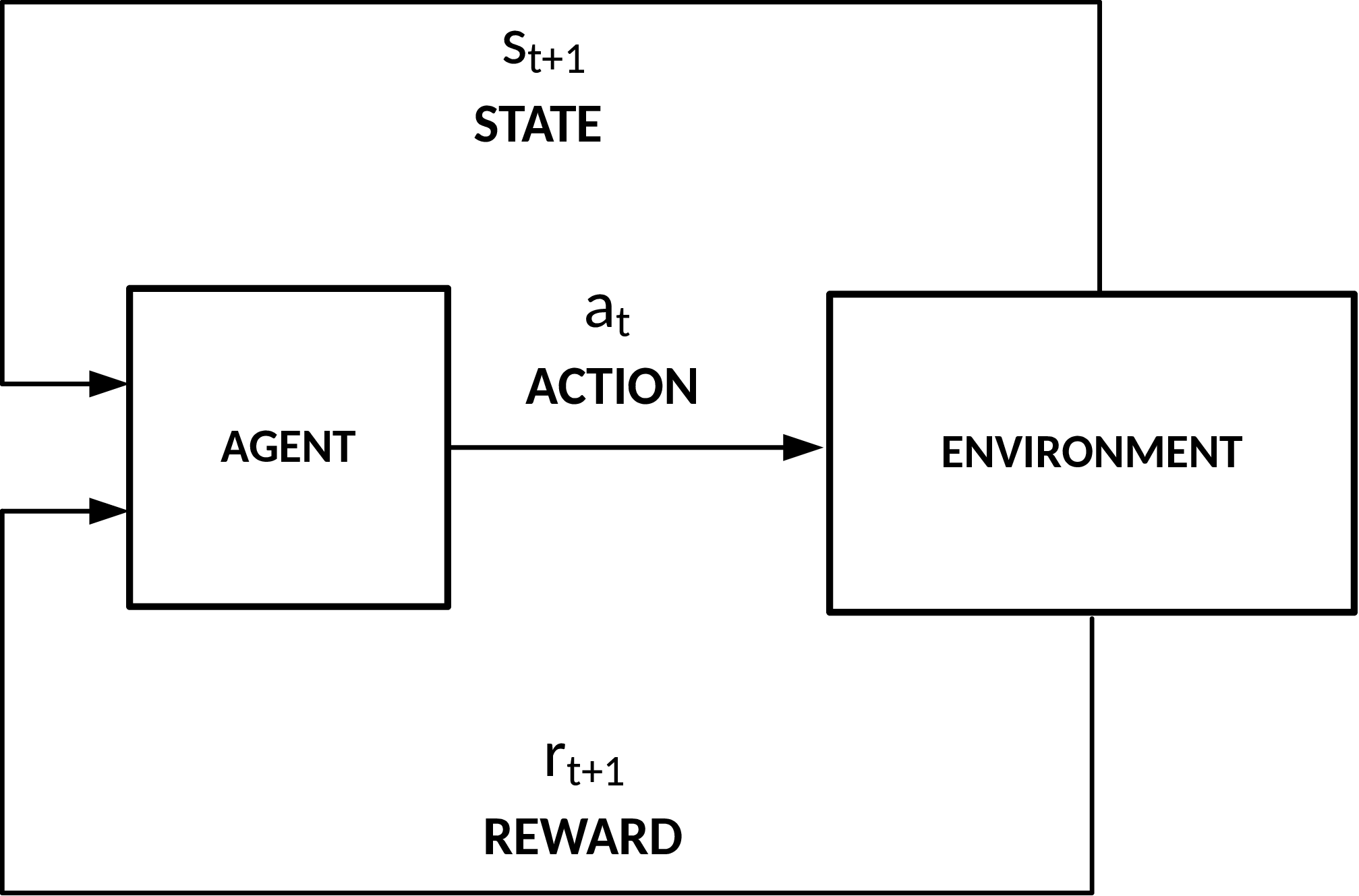}
	\caption{Interaction between the agent and the environment in an RL setting.}\label{fig1}
\end{figure}

\vspace{10pt}
Reinforcement Learning (RL) and its deep-learning variant (DRL) are used in many disciplines since they are powerful tools to solve complex problems that involve sequential decision-making and require extensive experience. Unlike supervised learning, in which an external intelligence labels the training data, or unsupervised learning, in which the goal is to find hidden patterns in unlabeled data, RL is based on learning, by trial and error, from the samples generated within the interaction of an agent and its environment. The general formulation of an RL problem is shown in Fig.~\ref{fig1}. In each step, the agent, which is the decision-maker, observes the current environment state and the reward obtained and, following a policy, decides which action to perform next. The agent's final goal is to maximize the cumulative reward, named expected return \cite{Sutton2018}.

RL algorithms can be classified into model-free or model-based methods. The main difference between them is that, in the former, the agent only learns by trial and error the policy that maximizes the objective, whereas in the latter, the agent can also employ a function that predicts the state transitions and rewards from the environment. Model-based algorithms \cite{Silver2017} allow the agent to plan ahead and decide among the best possible options, reducing the number of samples needed in a model-free approach and achieving an outstanding sample efficiency. Unfortunately, the ground-truth model of the environment is not always available. In those cases, the agent must learn the representation from experience. 

Robotic arms were previously limited to repetitive movements in a fixed setting. Nevertheless, nowadays, they can deal with stochasticity. As stated in \cite{Mahmood2018}, model-free approaches are bound to perform better than model-based solutions in scenarios where some components may be placed randomly in the workspace. As aforementioned, the main drawback of model-free RL algorithms is the amount of experience required to start learning. In a robotic arm, generating enough experience by trial and error could lead to several dangerous situations like collisions between the manipulator and its surroundings that can result in structural damage to expensive assets. Hence, being able to model and train in a physically realistic virtual environment is crucial to allow the agent to explore and exploit suboptimal policies that would have undesirable consequences in a real scenario. 

This approach fairly mitigates the problem associated with the low sample efficiency of DRL methods \cite{Lavet2018}, but causes a new issue: the efficient transference of the synthetic experience into the real world, broadly known as the sim-to-real problem.  

To address this problem, we draw on the work about progressive neural networks (PNNs) presented in \cite{Rusu2017}. In this architecture, a single big neural model trained with synthetic experience (the virtual column of the PNN) connected with as many light neural models as required to tackle real-world problems (the real columns of the PNN). These models in charge of interfacing the real world are trained with real experience. We regard this approach as very promising for two main reasons: 1) it allows to generate good representations of real situations from relatively few samples of real experience; and 2) it makes it possible to use the same set of synthetic data to help the same agent master different real tasks. 

However, to exploit the full potential of the PNN approach, training the agent appropriately in the virtual environment is crucial. This paper poses a straightforward research question: how robust a given agent is right after the virtual training phase (i.e., before modifying the weights of the virtual columns with the information provided by the real experience). Following our rationale, an agent which exhibits high robustness to discrepancies between the virtual environment and reality in this specific moment will require much less real experience to be fully operative than an agent which is highly sensitive to this effect. Therefore, we indirectly obtain a measurement of the amount of real experience DRL agents require to perform well by measuring robustness after the virtual training stage. 

To respond to this question, we obtained first some robustness results following the approach of the original PNNs paper \cite{Rusu2017}, in which the parameters of the virtual model were fixed. Then, we have compared this with an alternative approach where we introduced randomness in some particular virtual model parameters. With this idea, which falls within the field of Domain Randomization (DR) techniques \cite{Tobin2017}, we aimed to simulate, during training, the uncertainty that could later be translated into discrepancies between the virtual and the real worlds.  

The main contributions of this paper are: 1) an original benchmark to measure the robustness of any artificial agent that combines virtual and real experience, and 2) the quantification of the gain in robustness that is obtained by randomizing the virtual environment in the PNN approach, which can be interpreted as an increase in the efficiency of the sim-to-real process. With our results, we have also prepared the field for the hybridization of PNNs and DR, which, to the top of our knowledge, is something still to be explored.

The remainder of the paper has the following structure. Section~\ref{sec2} refreshes some basic reinforcement learning concepts, helpful in understanding the model's particular choices we present later. We recommend skipping this section if the reader is familiar with RL concepts. Section~\ref{sec3} analyzes the state-of-the-art of DRL in robotics, DR to address the differences within real and simulated worlds, and RL robustness analysis techniques. Section~\ref{sec4} describes the materials used in this research to support our experiments. Section~\ref{sec5} presents the followed method to answer the aforementioned research question. Section~\ref{sec6} exposes the results obtained and discusses the findings. Finally, in Section~\ref{sec7}, we report conclusions and future work. 

\section{Preliminaries}\label{sec2}
\vspace{10pt}
RL is a Machine Learning (ML) area in which an agent learns from the feedback given by the environment during the interplay between them. The elements that define an RL problem are the environment's state, which the agent perceives as an observation $S_t \in S$ where $S$ is the set of states; the action the agent performs $A_t \in A(S_t)$, being $A(S_t)$ the set of actions available in $S_t$; and the reward $R_{t+1} \in R \subset {\rm I\!R}$, which is the result of the agent's decision and the environment dynamics (Fig.~\ref{fig1}). Besides, the agent acts following a policy $\pi_t(a \mid s)$, which maps states to actions. The agent seeks to maximize the reward received over time, being this expressed as $G_t = \sum_{k=0}^{\infty} \gamma^kR_{t+k+1}$ where $\gamma \in [0, 1]$ is the discount rate that modulates the effect of future rewards in the present moment, $k$ denotes the number of time steps, and $t$ refers to the current time step. 

RL problems can be formulated as a Markov Decision Process (MDP) \cite{Bellman1957} or a Partially Observable Markov Decision Process (POMDP) \cite{Monahan1982}. However, since the mathematical tools available to solve POMDPs do not usually scale well, these problems must be converted to an MDP by means of a transformation function that allows obtaining the environment state from the observations \cite{Sutton2018}.

A finite MDP is a tuple $(S, A, P, R, \gamma)$ \cite{Silver2020} consisting of  a finite set of states $S$, a finite set of actions $A$, a state transition probability matrix $P$, a reward function $R$, and a discount factor $\gamma$. 

RL model-free algorithms can be classified into two main groups: value-based or policy-based methods \cite{Lavet2018}, \cite{Al-MasrurKhan2020}. Let's focus on the second category, to which the algorithm applied in this research belongs. Policy-based methods optimize the parametrized policy as $\pi_\theta(a \mid s)=Pr[a \mid s; \theta]$ by either using gradient ascent or maximizing local approximations on the expected return. While value-based strategies could be trained off-policy, the optimization in the policy-based algorithms is usually undertaken on-policy using the information of the experience obtained with the last version of the policy. 

\begin{figure}[h]
	\centering
	\renewcommand{\figurename}{Fig.}
	\includegraphics[width=0.3\textwidth]{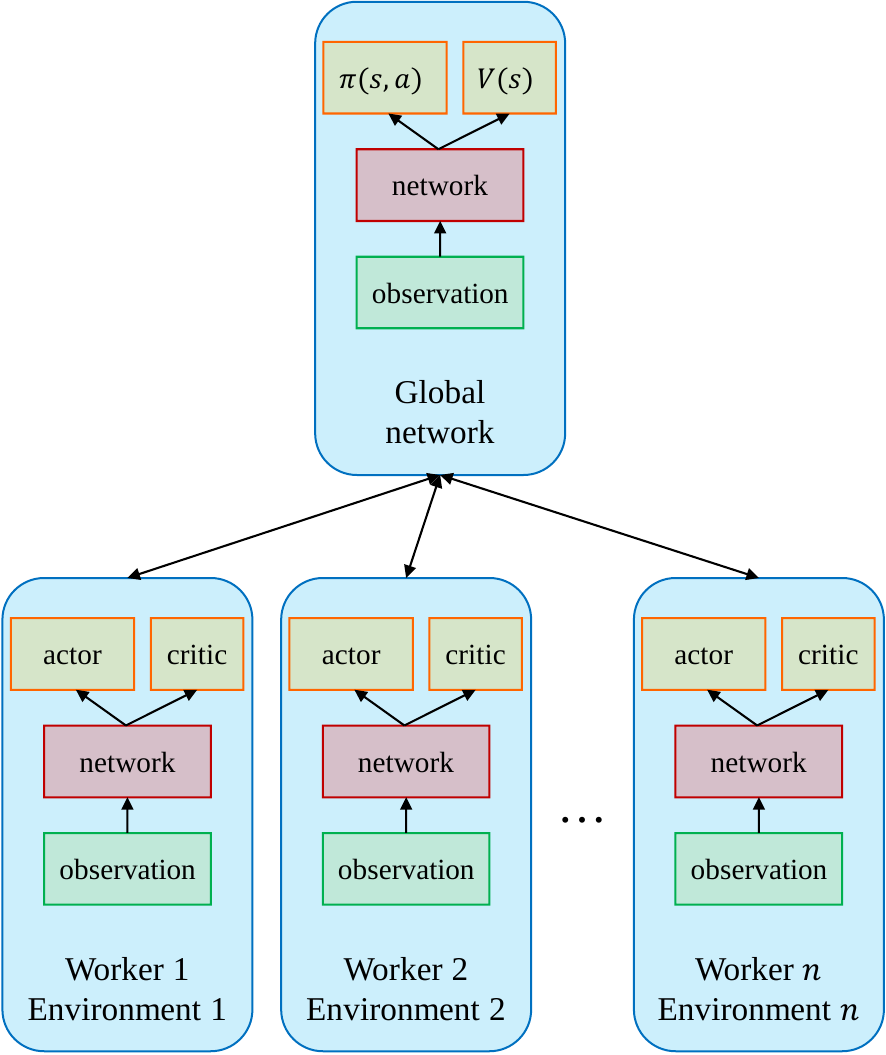}
	\\
	\caption{A3C general architecture diagram. Each worker trains and evaluates their networks individually until a certain number of steps in which the parameters are shared to the global network to update it with the ones that result in better outcomes.}\label{fig2}
\end{figure}

In the Asynchronous Actor-Critic (A2C) \cite{Mnih2016} and Asynchronous Advantage Actor-Critic (A3C) \cite{Mnih2016}, which we consider a policy-based method, even though it sometimes appears classified in the literature in the middle of both groups, the Actor oversees the policy optimization, while the Critic's role is to estimate the value function, that is the expected value for an agent to be in a specific state. The term asynchronous refers to the fact that the network has multiple agents that operate individually, interact independently with their environment and only share the results every $n$ steps to update the global network. The global scheme of an A2C and A3C is presented in Fig.~\ref{fig2}.

The A3C is explained in more detail in Appendix~\ref{secA1} since it is the algorithm implemented in this research due to its strengths in DRL problems with discrete action spaces and the successful results of Rusu et al. \cite{Rusu2017} in their experiments. 

\section{Related work}\label{sec3}
\vspace{10pt}
RL approaches have traditionally failed to scale to high dimensional spaces like the ones modern artificial agents face when they need to learn from camera images that monitor the environment \cite{Lazaridis2020}. DRL overcomes the limitations of RL methods like memory and computational complexity thanks to the function approximation and representation learning properties that deep neural networks exhibit \cite{Arulkumaran2017}. Therefore, the burst of DRL comes along with the growth of some deep learning algorithms such as Convolutional Neural Networks (CNNs) and Recurrent Neural Networks (RNNs) like Long-Short Term Memory (LSTM) networks \cite{Goodfellow2016}, \cite{Lecun2015}. These mechanisms enable a broad feature abstraction level that, in practice, eliminates the need for feature engineering. Besides, expanding these mechanisms to all the research community was possible thanks to the fast hardware development and its economic accessibility.

DRL provides the mechanisms to solve complex issues in robotics like optimal control policies in high dimensional spaces, the interaction with dynamic environments, high-level task planning, and advanced manipulation, among others \cite{Kober2013}. Though DRL facilitates robot task solutions with its representation generally based on the state-action discretization and Bellman equations approximations \cite{Bellman1957}, some challenges still exist. These hurdles might be under-modeling, which involves failing to achieve an accurate virtual scenario and model dynamics, triggering poor performance in the real world, the reward design, and the goal specification \cite{Li2019}.

Robot applications span a wide range of fields, from agriculture to cooking; however, most research in RL applied to robotics focuses on autonomous navigation, dexterity, and manipulation. For instance, in robot navigation, it is worth mentioning the work in \cite{Zhu2017}, where they use two images, one as the first-person view and the other of the target, as the A3C inputs. \cite{Zhang2017} enlarged a deep successor representation for the Q-value to deal with new environments by transferring the policies. In \cite{Ejaz2021}, the authors propose a network architecture based on layer normalization and Dueling Double DQN to learn how to avoid collisions using four consecutive depth images as inputs. After being trained in a simulated environment, the implementation into the real world could be done directly. Finally, some analyses also fuse Simultaneous Localization and Mapping (SLAM) \cite{Durrant-Whyte2006}, \cite{Bailey2006} with DRL networks. For example, \cite{Wen2020} suggests combining SLAM and path planning. They implement the Dueling DQN algorithm to avoid obstacles at the same time as FastSLAM is building the 2D environment map. For further research examples in navigation and manipulation applying DRL, refer to \cite{Tai2016}.  

DRL applied to robots' dexterity and manipulation concerns tasks like reaching a certain point in the workspace or picking objects and placing them elsewhere. As defined in \cite{Li2019}, manipulation tasks are generally determined by: 1) collecting experience, 2) training a policy and, depending on the application, 3) predicting the object pose, and 4) combining the previous information to follow a trajectory and grasp an object. One of the most recognized publications in dexterous robotics is Dexterity Network (Dex-Net). It is an open-source project developed by the Berkeley Automation Lab that designed a complete algorithm set. It can identify different devices in front of the robot, plan the robot trajectory to each object's best grasping point, and clamp with the gripper or suck with the vacuum cap, whatever is most suitable \cite{Mahler2016}, \cite{Mahler2017_1}, \cite{Mahler2018}, \cite{Mahler2017_2}, and \cite{Mahler2019}. On the other hand, to improve grasping tasks \cite{Kalashnikov2018} introduce a scalable self-supervised vision-based RL framework designed upon a generalization of the Q-learning method applied to continuous action sets. One of the main drawbacks in this research is that their method implies the collection of real grasping data, a task that is time-consuming and, in most cases, unaffordable since the number of available assets is limited.  

The real data gathering problem can be solved using a virtual setup in which synthetic experience is generated and utilized to learn optimal policies. This sim-to-real approach opens new dilemmas like how to transfer the knowledge to the real environment, which might differ from the virtual one. Several transfer learning techniques have been tested to bridge this gap, which is even more relevant in DRL with visual inputs. The simplest method, fine-tuning, has revealed that the model adjustment process might be tedious for complex tasks, and the results are not always the expected. One of the state-of-the-art transfer learning systems is Domain Adaptation (DA), which succinctly translates images from a source domain to a target domain. \cite{Shoeleh2020} presents an approach applicable to RL problems with continuous action spaces that speed up learning new tasks using DA. Generative Adversarial Networks (GANs) \cite{Goodfellow2014} are widely used as a DA tool to either adapt images from a virtual environment to make them look more like the real scenario or to modify somehow the characteristics of the environment elements as shown in \cite{Taigman2016}, \cite{James2019}, \cite{Shrivastava2017}, \cite{Zhu2017_2} and \cite{Gamrian2020}. Another technique that can be implemented individually \cite{Zhu2018} or along with GANs \cite{Ho2016} is Imitation Learning (IL), where the policy is learned from expert demonstrations made by an agent that has correctly learned the task before. 

On the other hand, an encouraging method to transfer the knowledge learned in a virtual setup was presented by Rusu et al. \cite{Rusu2017} in their work named ``Sim-to-Real Robot Learning from Pixels with Progressive Nets.'' This investigation was based on their previous research, ``Progressive Neural Networks'' \cite{Rusu2016}, whose object was to avoid catastrophic forgetting in sequences of tasks. Instead of relying on the render engine of the virtual environment, or the image processing needed in the previously mentioned techniques, that might involve high computational costs, \cite{Rusu2017} suggests that an agent could be trained in simulation and, through column's lateral connections, being a column the network trained in a virtual or real domain, the knowledge learned can be transferred to the real world. Hence, the model obtained in the virtual scenario can be just slightly adapted to behave appropriately in the real environment. Moreover, the reuse of all the experience learned during training could be successfully transferred to accomplish new complex tasks. 

In the experiments presented in \cite{Rusu2017}, the agent's objective was to reach a certain point in the workspace with the grippers, first in simulation and then in the real domain. The robotic arm, a Jaco, was modeled in MuJoCo \cite{Todorov2012}, and the input to the DRL algorithm, an A3C, was the raw $64x64$ RGB image. The outputs were the nine discrete joint policies and the value function, in total, 28 values. They performed different experiments varying the columns' network architecture and size, reaching a higher reward with the recurrent network than with the feedforward architecture, achieving a 100\% success in the real scenario with a sparse reward configuration. Due to its results and applications, the model described in \cite{Rusu2017} was the inspiring framework adopted in this paper.

Nevertheless, to minimize the train needed for the real column, our research aimed to make a preliminary robustness analysis of the policy model subject to different camera positions and orientations and, depending on the outcomes, improve the virtual model including more diversity during its training. The changes and modifications made between the Baseline model presented in this document and the proposal by Rusu et al. are explained in Appendix~\ref{secA1}. 

Domain Randomization (DR) is a DA technique where training is conducted in a virtual domain whose attributes can change according to what might happen in a real scenario. As demonstrated in \cite{Tobin2017}, training models in simulated environments with enough variability could approximate real scenes, making the agent agnostic to the workspace. They trained a modified VGG-16 CNN \cite{Simonyan2015} detector with rendered images generated in random camera positions, lighting conditions, object positions, and textures to mimic their physical setup. Without further training or parameter adjustment, they achieved a \SI{1.5}{cm} accuracy in the real world.

The results from \cite{Jakobi1995} showed that the rift from simulation to the real environment could be handled by introducing noise to the simulation training. The robot performance was enhanced by implementing some disturbances to the training. More recently, Mozifian et al. \cite{Mozifian2020} exhibited that applying domain randomization can also reduce the reality gap in robotic manipulation. By adding random perturbations that were not realistic and implementing their Intervention-based Invariant Transfer learning (IBIT) method, they improved the model generalization from simulation to reality. 

Concerning the agent's robustness, some publications like \cite{Chan2020} and \cite{Jordan2020} suggest methods to evaluate the model performance and shape robustness during and after training. In our research, we have considered the problem as a classification dilemma where the episode could be labeled as success or failure if the goal was reached or not. Hence, the metrics used to analyze and compare the trained models were the learning time needed, the average accuracy obtained, and the maximum failure distance in evaluation.  

\section{Materials}
\label{sec4}
\vspace{10pt}
The experiments performed in this research make use of a set of virtual and physical elements listed in this section.

\textbf{\underline{Deep-learning computer}}

The hardware used to perform the experiments was a PC running Ubuntu 20.04 equipped with an Intel Core i9-10900KF processor at \SI{3.70}{GHz}, \SI{64}{GB} of DDR4 RAM, an NVIDIA GeForce RTX 2080 Ti GPU, and a \SI{2}{TB} M.2 SSD. The algorithms are programmed in Python 3.8, using PyTorch \cite{Stevens2020} to implement the advanced neural model in the core of the artificial agents and other libraries for basic computation, ancillary graphic services, etc. (e.g., NumPy, Matplotlib).

\textbf{\underline{Virtual Environment}}\nopagebreak

All the experiments presented in this paper are performed in a virtual setup made up of the robotic arm IRB120 from ABB\footnote{https://new.abb.com/products/robotics/industrial-robots/irb-120/irb-120-data}, a fixed externally-mounted monocular camera, and a red \SI{0.03}{m} cube (target) the robot arm must reach in a maximum number of steps (an episode). The IRB 120 has 6 degrees of freedom (DoF), a reach of \SI{0.58}{m}, a payload of \SI{3}{kg}, and an armload of \SI{0.3}{kg}. Its operating ranges and velocities are included in Table~\ref{tab1}.
\begin{table}[h!]
        \centering
        \caption[IRB120 joints' working range and velocity]{IRB120 joints' working range and velocity.}
        \addtolength{\tabcolsep}{4pt}
        \renewcommand{\arraystretch}{1.5}
        \resizebox{0.6\textwidth}{!}{%
        \begin{tabular}{lcc}
            \toprule
            \textbf{Axis motion} & \textbf{Working range} & \textbf{Velocity} \\
            \midrule
            Axis 1 (rotation) & -165 to \SI{+165}{\degree} & $\pm$\SI{250}{\degree/\second} \\
            Axis 2 (arm)      & -110 to \SI{+110}{\degree} & $\pm$\SI{250}{\degree/\second} \\
            Axis 3 (arm)      & -110 to \SI{+70}{\degree}  & $\pm$\SI{250}{\degree/\second} \\
            Axis 4 (wrist)    & -160 to \SI{+160}{\degree} & $\pm$\SI{320}{\degree/\second} \\
            Axis 5 (bend)     & -120 to \SI{+120}{\degree} & $\pm$\SI{320}{\degree/\second} \\
            Axis 6 (turn)     & -400 to \SI{+400}{\degree} & $\pm$\SI{420}{\degree/\second} \\
            \bottomrule
        \end{tabular}
        }
        \label{tab1}
\end{table}

\begin{figure}[h]
	\centering
	\renewcommand{\figurename}{Fig.}
	\subfloat[\centering]{{\includegraphics[width=0.2\textwidth]{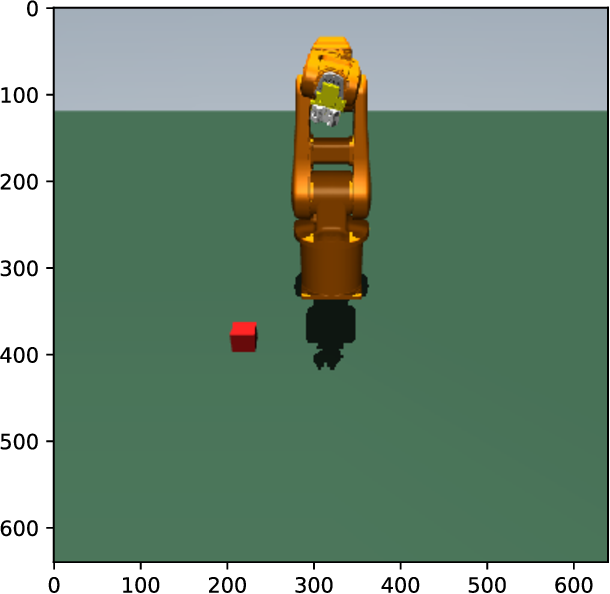} }}
	\hfill
	\subfloat[\centering]{{\includegraphics[width=0.21\textwidth]{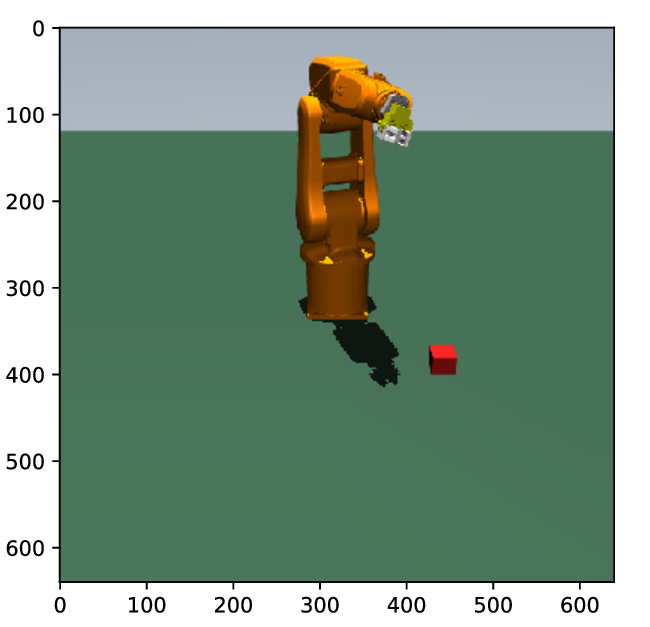} }}
	\hfill
	\subfloat[\centering]{{\includegraphics[width=0.21\textwidth]{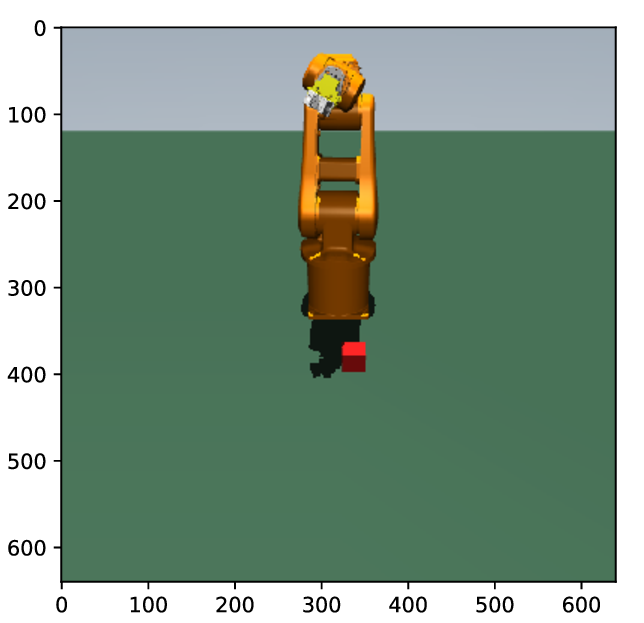} }}
	\caption{Random initial positions of the target and joints being the camera located at $(180^{\circ}, -30^{\circ})$. Axes in the three pictures indicate pixel coordinates.}\label{fig3}
\end{figure}

The robot's behavior has been simulated in MuJoCo to capture its physics accurately. Since rich visual textures were not considered at this research stage, the virtual environment was simply rendered using Matplotlib. Fig.~\ref{fig3} presents examples of the environment image captured by the camera. The target (red cube) and the initial joint's position were placed randomly in the workspace area. 

\section{Method}\label{sec5}
\vspace{10pt}
This section presents the general details of the research approach followed, the model design and training details, the virtual test bench used to perform robustness test cycles, and the analyses performed after them. 

\textbf{\underline{Research approach}}\nopagebreak

\begin{figure}[h]
	\centering
	\renewcommand{\figurename}{Fig.}
	\includegraphics[width=\textwidth]{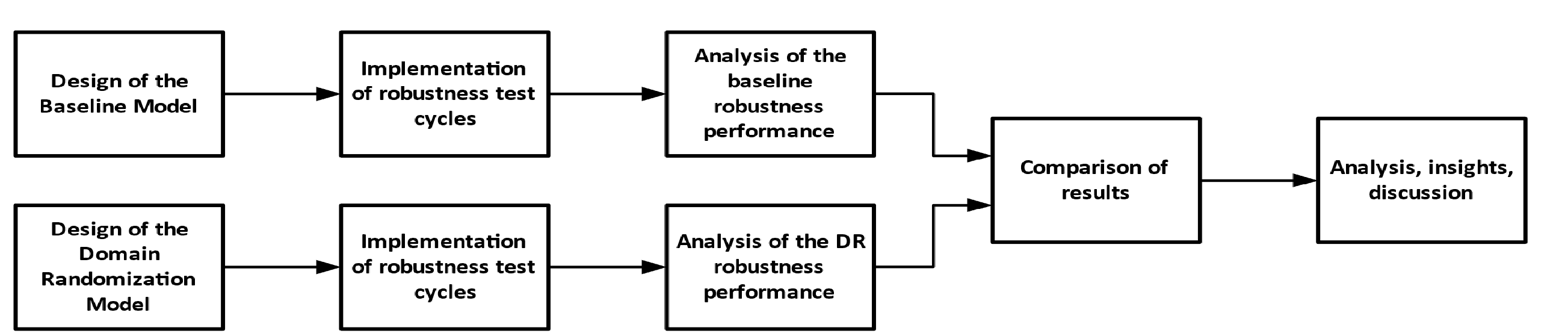}
	\caption{Schematic diagram of the methodology.}\label{fig4}
\end{figure}

Fig.~\ref{fig4} summarizes the procedure carried out in this research. On the top branch, there is an artificial agent to control the robot following the PNN approach found in the literature. This is the reference model, coined Baseline model (BM) or Baseline agent, that allows quantifying the improvement introduced by the proposed approach. Then, a virtual test bench measures the robustness of this BM to changes in the camera setting, and finally, the results are structured and analyzed.

Afterward, shown on the bottom branch, the same method is followed to obtain the robustness results of an improved artificial agent in which the training phase is enriched by adding variability, that is, applying DR to the camera position. For the sake of simplicity, only the camera position is used to represent discrepancies between the virtual and the real worlds, acknowledging that the model is sensitive enough to changes in this variable. We refer to this model as the DR model (DRM) or DR agent. 

The results and insights extracted from comparing the outcomes of both models are presented in Section~\ref{sec6}.

\textbf{\underline{Training and model design details}}\nopagebreak

Agent training is made up of episodes. At the beginning of each one, the target and robot positions are initialized randomly according to a uniform distribution whose limits are the working area for the target and the joints' degrees that prevent the robot position from exceeding the image limits, that is, a 15\% of its operating range (Table~\ref{tab1}). An episode ends when the number of steps reaches 50 or if the distance from the gripper to the target is lower or equal to \SI{5}{cm}. 

The discount factor ($\gamma$) is fixed to 0.99 in the experiments since the agent should evaluate each action regarding the future rewards in this problem. The experience presented to the agent during training amounts to 70 million steps. Every 50 thousand steps, training is halted to perform 40 evaluation episodes and compute the average, maximum, and minimum distance from the gripper to the target position, obtaining the training curves that allow measuring the effectiveness of training, sample efficiency, etc. 

As aforementioned, the BM is designed following a large extent of the PNN approach. However, an appreciable effort has also been put into studying the effect of the MDP's action space and the design of the reward signal to acquire a better understanding of the drivers of control performance. Although we believe this part of the work is interesting, we consider it out of the main scope of the paper and, therefore, the details of the BM design, including the choice of the algorithm's hyperparameters, have been relegated to Appendix~\ref{secA1}. 

The most important detail about the BM to properly follow the robustness results in this paper is that the camera position is fixed during training in both axes: $180^\circ$ for the $z$-axis and $-30^\circ$ for the $y$-axis, a setup that provides a close to optimal perspective of the robot and the target positions, as observed in Fig.~\ref{fig3}. While the DRM inherits the action space, reward signal, and hyperparameters of the BM, it also includes variability in the camera position on both axes during training. Specifically, the $z$-axis camera position might vary in the interval $[160^\circ, 200^\circ]$, and the $y$-axis, in $[-40^\circ, -20^\circ]$.

Table~\ref{tab2} summarizes the details of the training parameters of both models. WRIL stands for Working Range Inferior Limit and WRSL for Working Range Superior Limit (Table~\ref{tab1}), $U(x, y)$ means this parameter was obtained by sampling a uniform distribution from $x$ to $y$. To ensure a fair comparison between models, all parameters are identical except for the randomized variables in the proposed DR approach (those corresponding to the camera location). 
\begin{table}[h]
	\renewcommand{\thefootnote}{\alph{footnote}}
	\begin{center}
		\begin{minipage}{\textwidth}
		\caption{Training parameters used for the Baseline and the DR models.}
		\label{tab2}
		\renewcommand{\arraystretch}{1.8}
		\resizebox{\textwidth}{!}{
			\begin{tabular}{cC{1.2cm}C{1.4cm}C{2.0cm}C{2.2cm}C{2.0cm}C{2.0cm}}
				Model & Number of steps & Success distance\footnote{The success distance is measured from the robot's gripper to the target position.} & Episode initial position\footnote{The robot's initial position is set according to a uniform distribution in which minimum and maximum value are respectively a 15\% more and a 15\% less of the robot's working range inferior and superior limit.} & Episode target position\footnote{The episode target position is also sampled from two uniform distributions, one for the $x$-coordinate and the other for the $y$-coordinate, whose boundaries represent the distance in cm from the robot base.} & Episode camera $z$-axis & Episode camera $y$-axis \\
				\hline
				\hline
				Baseline & 70 million & \SI{5}{cm} & $U(+15\% WRIL, \linebreak -15\% WRSL)$ & $U(0.2, 0.4)$ cm - $x$ coordinate \linebreak \linebreak $U(-0,3, 0.3)$ cm - $y$ coordinate & $180^\circ$ & $-30^\circ$ \\
				\hline
				DR & 70 million & \SI{5}{cm} & $U(+15\% WRIL, \linebreak -15\% WRSL)$  & $U(0.2, 0.4)$ cm - $x$ coordinate \linebreak \linebreak $U(-0,3, 0.3)$ cm - $y$ coordinate & $U(160^\circ, 200^\circ)$ & $U(-40^\circ, -20^\circ)$\\
				\hline
			\end{tabular}
		}
		\end{minipage}
	\end{center}
\end{table}

\textbf{\underline{Evaluation details: the virtual test bench}}\nopagebreak

After the design and training phases, the control robustness is methodically measured by assessing the performance of the given model when it only uses the virtual column (i.e., right before including real experience into the lateral connections and real columns of the PNN architecture). Our test bench allows changing almost all parameters in the virtual environment. To make the training-test pipeline clearer: the agent is first trained with a virtual model; then, the agent's policy is extracted from the neural network and tested against a different virtual model of the environment, which acts as a surrogate of the real model. This testing approach provides a highly interpretable measurement of the effectiveness of the virtual training stage of the agent.  

\begin{figure}[h]
	\centering
	\renewcommand{\figurename}{Fig.}
	\includegraphics[width=0.9\textwidth]{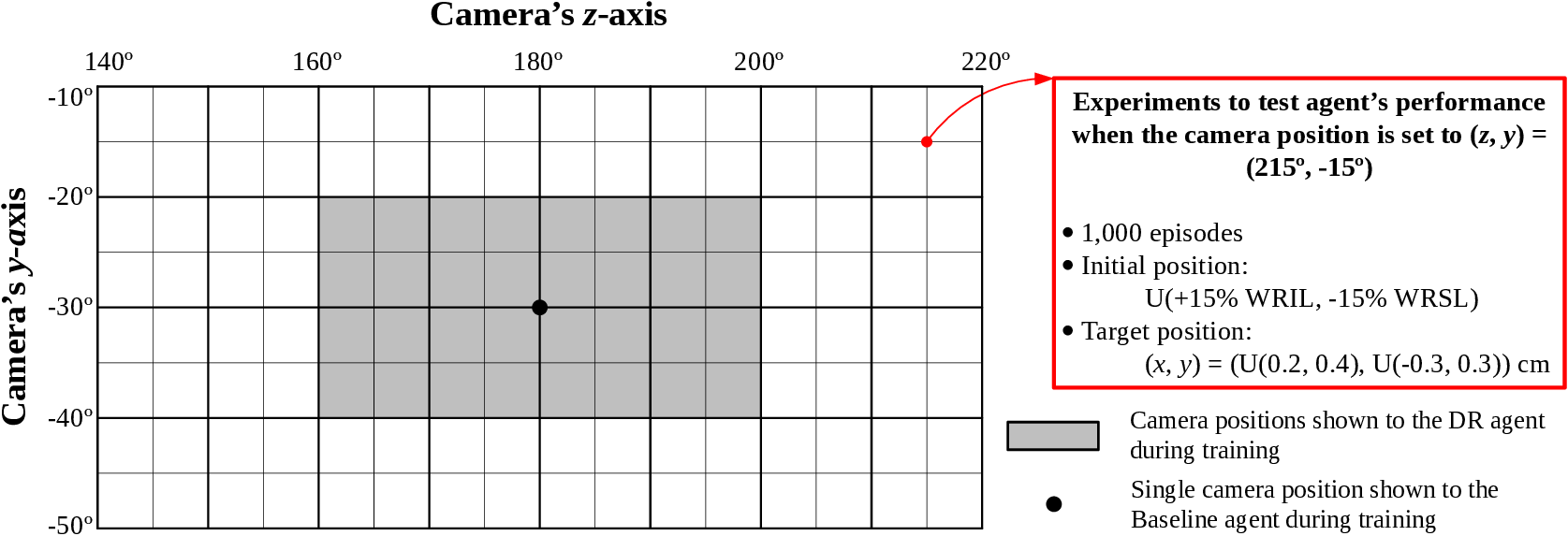}
	\caption{The virtual test bench grid designed to evaluate the models. The black dot refers to the camera's position in the BM training and the grey area to the camera orientations presented to the DR agent. The grid granularity is $5^\circ$.}\label{fig5}
\end{figure}

During the evaluation of both models, the episode is successfully completed if the distance between the gripper and the target is \SI{10}{cm} or less. Note that the tolerance was increased in \SI{5}{cm}, trusting that that training the agent in more challenging conditions than the ones actually required in the evaluation would lead to a better performance. In order to prevent over-optimistic results, the virtual test bench has a higher variation in the camera position variables than the introduced in the DRM's training. The rest of the parameters of the training and test models remained unchanged. Fig.~\ref{fig5} shows the grid of camera positions assessed in each robustness test cycle. The camera position sweeps within the interval $[140^\circ, 220^\circ]$ for the $z$-axis and $[-50^\circ, -10^\circ]$ for the $y$-axis. The granularity is set to $5^\circ$ in both cases, yielding 153 camera positions. For every camera location, 1,000 different episodes are run, each with its own random set of initial and target positions. These positions are obtained by sampling two uniform distributions with the parameters shown in Table~\ref{tab2}. In total, 153,000 combinations of camera locations and initial and target positions are evaluated in every test cycle. To evaluate the models' robustness, the number of steps needed to meet the steady-stable regime during learning, the cumulative reward, the average accuracy, and the failure distance on the episodes where the target is not reached, are analyzed. These variables are defined in Table~\ref{tab3}.

\begin{table}[h]
	\renewcommand{\thefootnote}{\alph{footnote}}
	\begin{center}
		\begin{minipage}{\textwidth}
		\caption{Evaluation parameters definition.}
		\label{tab3}
		\renewcommand{\arraystretch}{1.8}
		\resizebox{\textwidth}{!}{
			\begin{tabular}{C{2.2cm}C{2.2cm}C{2.2cm}C{2.2cm}}
				Number of steps & Cumulative reward\footnote{$\gamma \in [0, 1]$ = discount rate; $k$ = number of time steps; $t$ = current time step.} & Average accuracy\footnote{As aforementioned, we consider this problem as a classification dilemma, where episodes are tagged as success ($S_{Ep}$) if the target is reached, or not success ($NS_{Ep}$) if it is not.} & Failure distance\footnote{Only defined for non-success episodes.}\\
				\hline
				\hline
				Number of steps needed to meet the steady-stable regime during learning & $G_t = \sum_{k=0}^{\infty} \gamma^kR_{t+k+1}$ & $\frac{S_{Ep}}{S_{Ep}+NS_{Ep}}*100$ & Relative distance (cm) between the gripper and the target \\
				\hline
			\end{tabular}
		}
		\end{minipage}
	\end{center}
\end{table}

The shadowed region in Fig.~\ref{fig5} represents camera positions that have been presented to the DR agent during training (of course, for the Baseline, the camera position is always at the center of the grid). As can be observed, the surface of positions not shown to the agent during training is three times the gray area. This parametrization allows measuring how the agent performs when it is interpolating within its experience and when it is extrapolating. 

\begin{figure}[h]
	\centering
	\renewcommand{\figurename}{Fig.}
	\subfloat[\centering]{{\includegraphics[width=0.32\textwidth]{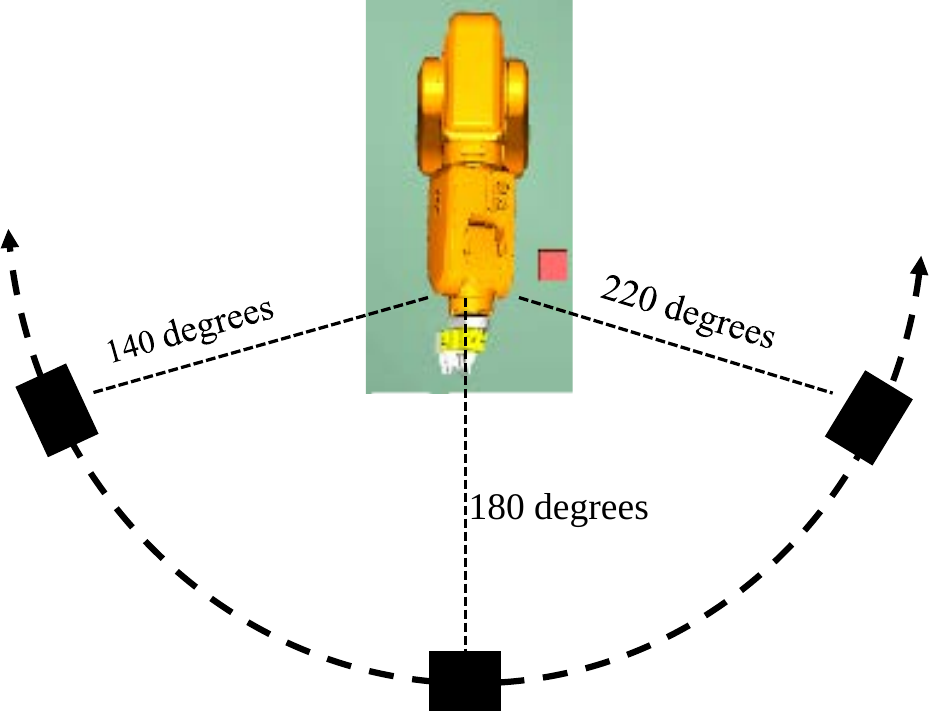} }}
	\hfil
	\subfloat[\centering]{{\includegraphics[width=0.22\textwidth]{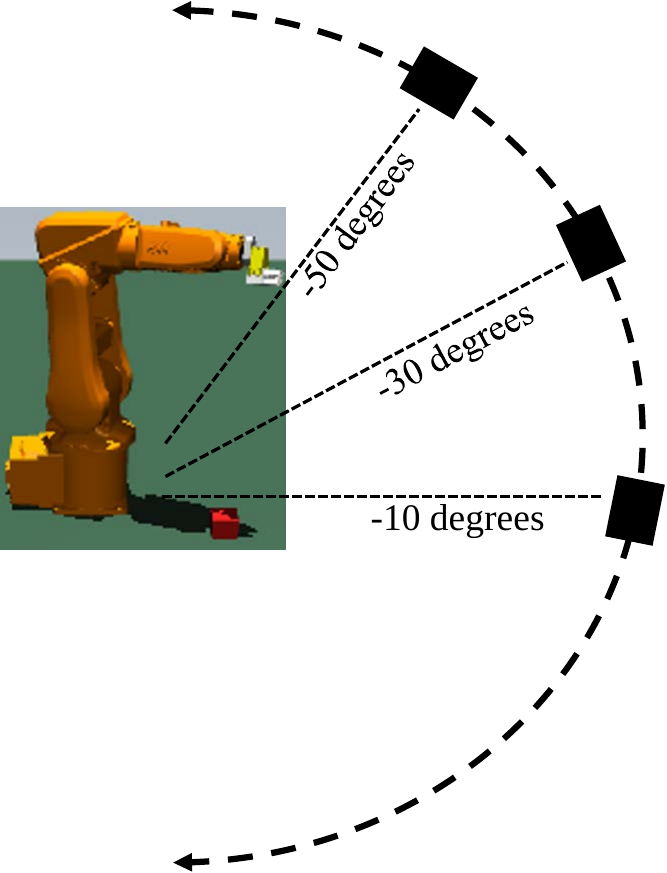} }}
	\caption{Range of possible camera orientations shown in evaluation around the $z$-axis (a) and $y$-axis (b), keeping the sphere radius $r = 2.0 m$ constant.}\label{fig6}
\end{figure}

Fig.~\ref{fig6} shows the axes and the limit positions of the camera to illustrate the viewpoint changes when it is moved along each axis.

\section{Results and discussion}\label{sec6}
\vspace{10pt}
This section presents the BM results first, followed by the DRM right after, repeating the same fixed scheme for both models to enhance understanding. In each of these parts, the training and the robustness 
\vspace{10pt}
\subsection{Baseline Model (BM)}\label{subsec6_1}
\vspace{10pt}
\textbf{\underline{Training}}\nopagebreak

\begin{figure}[h]
	\centering
	\renewcommand{\figurename}{Fig.}
	\includegraphics[width=0.6\textwidth]{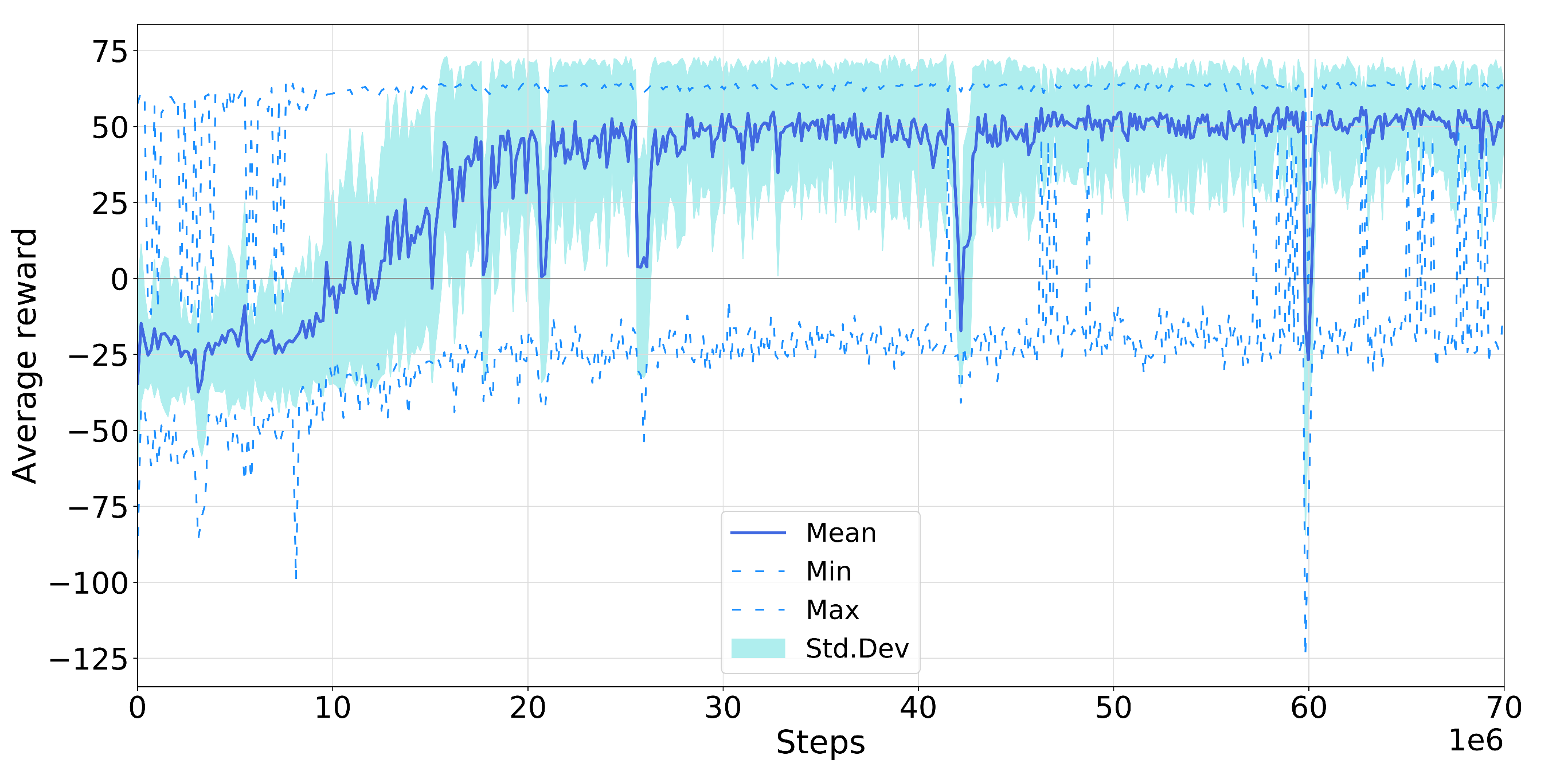}
	\caption{BM training curve across 70 million steps. The steady-stable regime, with an average reward of roughly 50, begins at around 35 million steps. Note that within this period the agent exploited two sub-optimal policies at steps 42 million and 59 million.}\label{fig7}
\end{figure}

Fig.~\ref{fig7} presents the training curve for the BM. The learning process meets the steady-stable regime in around 35 million steps. Note that around 40 million steps and 60 million steps, the agent seems to explore and exploit a sub-optimal policy, leading to a temporary decrease in the average reward. Under the conditions mentioned in Section~\ref{sec4} and Section~\ref{sec5}, the training time was 14 hours. 

\textbf{\underline{Robustness}}\nopagebreak

\begin{figure}[t]
	\centering
	\renewcommand{\figurename}{Fig.}
	\includegraphics[width=0.7\textwidth]{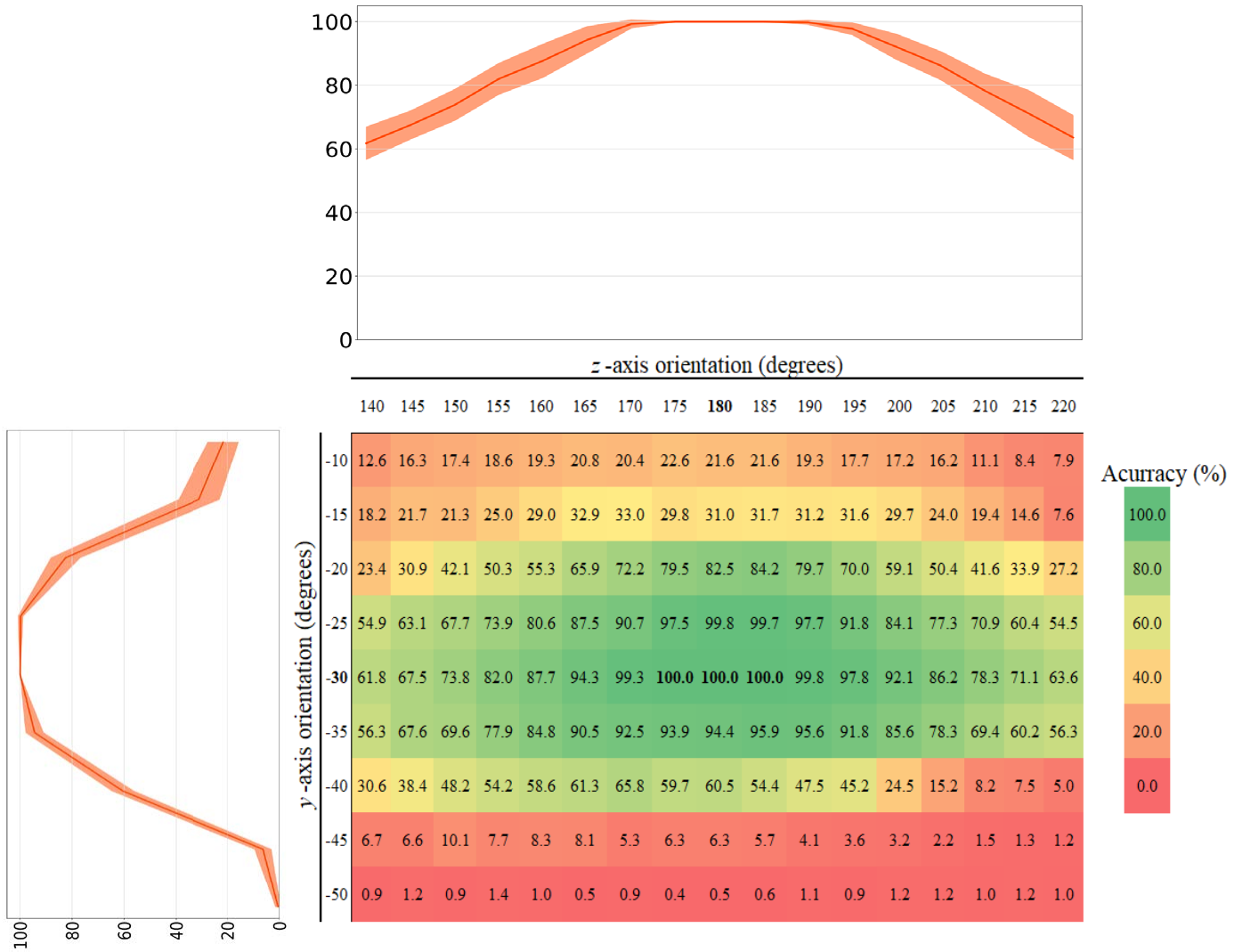}
	\caption{BM average accuracy obtained in evaluation presented as a heat map. The percentage shown for each camera pose combination is the average of 1,000 test episodes. The orientation highlighted in bold in the axes correspond to the position of the camera during the BM agent training. \linebreak On the top, the average accuracy achieved in evaluation across the $z$-axis is presented. It represents the projection of the heat map values when this one is intersected by a perpendicular plane to the $y$-axis at $-30^\circ$. \linebreak On the left, the average accuracy drawn in evaluation across the $y$-axis is showed. It represents the projection of the heat map values when this one is intersected by a perpendicular plane to the $z$-axis at $180^\circ$. \linebreak The line's shading corresponds to $\pm1$ standard deviation of the accuracy distribution per orientation pair.}\label{fig8}
\end{figure}

To better understand the accuracy results, the outcomes have been analyzed using a figure inspired in the orthographic projection system (Fig.~\ref{fig8}). The matrix, which corresponds to the floor view, is a heat map that presents the average accuracy obtained after evaluating the BM across all combinations of the $z$- and the $y$-axes. On top of it, the average accuracy is displayed along the $z$-axis when the $y$-axis orientation is fixed to $-30^\circ$, emulating a front view where the heat map surface intersects with a plane perpendicular to the $y$-axis at $-30^\circ$. Finally, the left plot shows the average accuracy when the $z$-axis is kept at $180^\circ$. This graphic can also be regarded as the surface projection that results from the intersection between the heat map and a plane perpendicular to the $z$-axis at $180^\circ$.

Although the Baseline agent was trained with a fixed camera position at $(180^\circ, -30^\circ)$, it keeps the accuracy above 90\% for the interval [$165^\circ, 195^\circ]$ in the $z$-axis and $[-35^\circ, -25^\circ]$ in the $y$-axis. It achieves 100\% accuracy at $(175^\circ, -30^\circ)$, $(180^\circ, -30^\circ)$, and $(185^\circ, -30^\circ)$, all of them with the height perspective in which the robot was trained. The maximum failure distance was \SI{48}{cm}, registered at $(140^\circ, -25^\circ)$ and at $(220^\circ, -50^\circ)$. Both positions correspond to unfavorable scenarios, where the camera is visibly below and over the initial $y$-axis orientation. 

Accuracy around the $z$-axis is distributed almost symmetrically for all the combinations from $-35^\circ$ to $-20^\circ$ in the $y$-axis. However, the results obtained for $[-50^\circ, -40^\circ]$ show that there is a divergence in the accuracy reached at $[140^\circ, 170^\circ]$ with respect to the ones in $[190^\circ, 220^\circ]$, especially at $-40^\circ$. We ponder that this behavior is caused by the effect of the robot's shadow in the picture captured when the camera was clearly above $-30^\circ$. This detail might increase in the difficulty of analyzing the input image since the contrast between the target's and the floor's colors, both affected by the shade, is reduced. This phenomenon produced a substantial decrease in the agent's performance. The same incident happened at $-15^\circ$ and $-10^\circ$ for the intervals $[140^\circ, 150^\circ]$ when the outcomes are compared to those within $[210^\circ, 220^\circ]$. Apparently, the agent analyzes differently two pictures with the same camera height, but taken from the robot's left side, where the shadow might not influence, and the right side, where the shade seems to be an important disturbance.

Regarding the results for the $y$-axis, Fig.~\ref{fig8} shows that the agent's behavior is not symmetrical either. The performance for the worst scenarios is considerably better when the camera is positioned below the robot's middle plane at $-10^\circ$ than when it is orientated above it at $-50^\circ$. Fig.~\ref{fig6} suggests that the agent might struggle to properly infer the joints' position at $-50^\circ$ since the camera perspective, combined with the robot position, could hide some robot links. However, at $-10^\circ$, the challenge is to deduce the depth because the camera orientation is nearly parallel to the floor plane. According to this, it is easier for the agent to manage the lack of depth data than the loss of joints' orientation information.

\subsection{Domain Randomization Model (DRM)}\label{subsec6_2}

\textbf{\underline{Training}}\nopagebreak

\begin{figure}[h]
	\centering
	\renewcommand{\figurename}{Fig.}
	\includegraphics[width=0.6\textwidth]{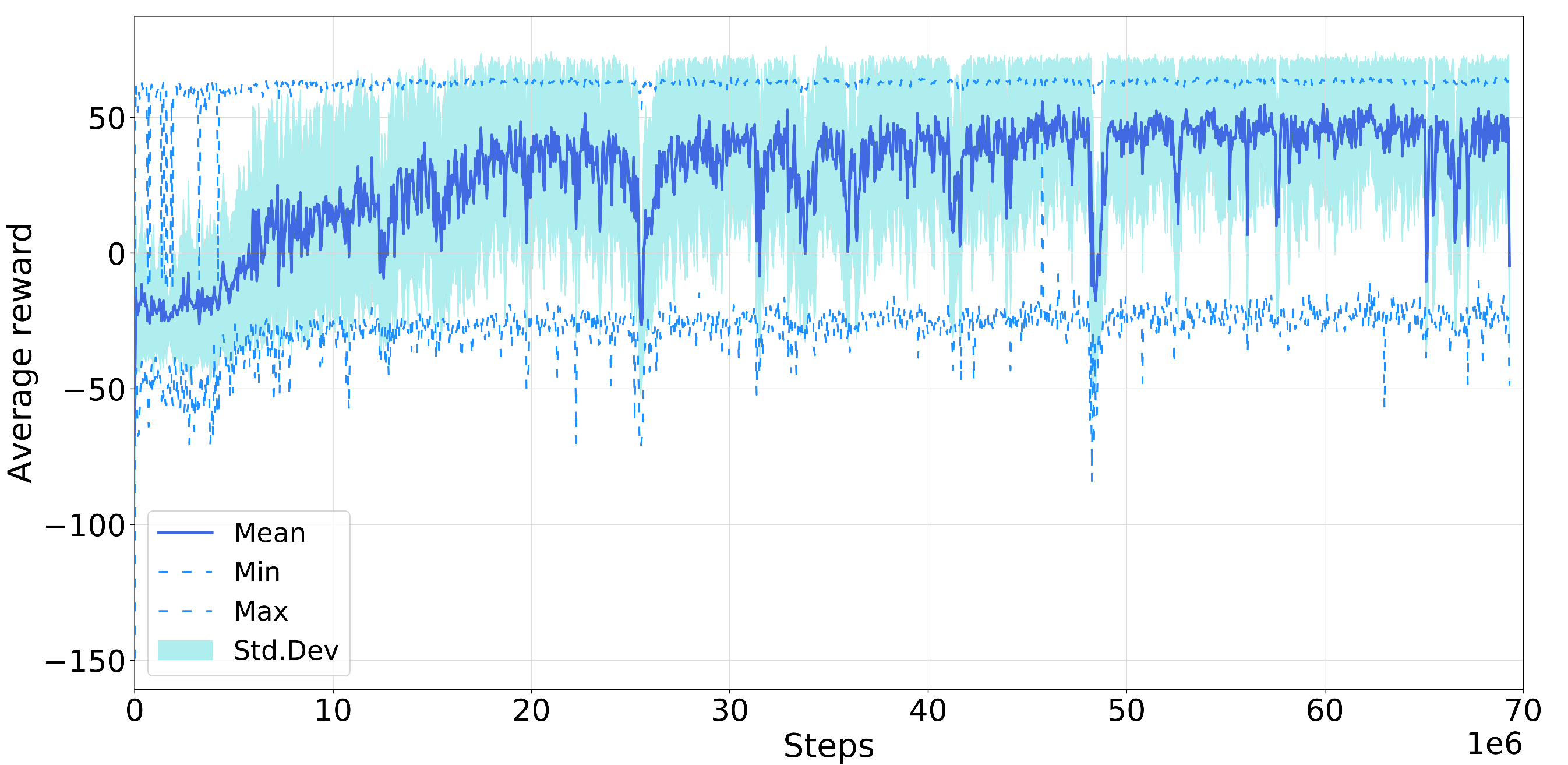}
	\caption{DRM training curve across 70 million steps. The steady-stable regime, with an average reward of 40, begins around 40 million steps. Note that the DR learning curve is more irregular than the one obtained for the BM, likely due to the noise and the problem complexity. In this case, the valleys caused by the sub-optimal policies are registered at 50 million and 65 million steps.}\label{fig9}
\end{figure}

Fig.~\ref{fig9} displays the training curve for the DRM. In this case, the steady-stable regime is reached at around 40 million steps, which is consistent with the increase of the problem complexity. Nevertheless, during the learning process, the agent exploited more sub-optimal policies than in the BM, the most relevant the ones in the steady-stable regime reached at 50 million and 65 million steps approximately. Furthermore, it is interesting to point out that the training curve is more irregular than the BM sequence shown in Fig.~\ref{fig7}. We consider this could be due to the noise and the training complexity. Keeping the same training conditions that in the BM, the average training time was 15 hours.

\textbf{\underline{Robustness}}\nopagebreak

\begin{figure}[h]
	\centering
	\renewcommand{\figurename}{Fig.}
	\includegraphics[width=0.7\textwidth]{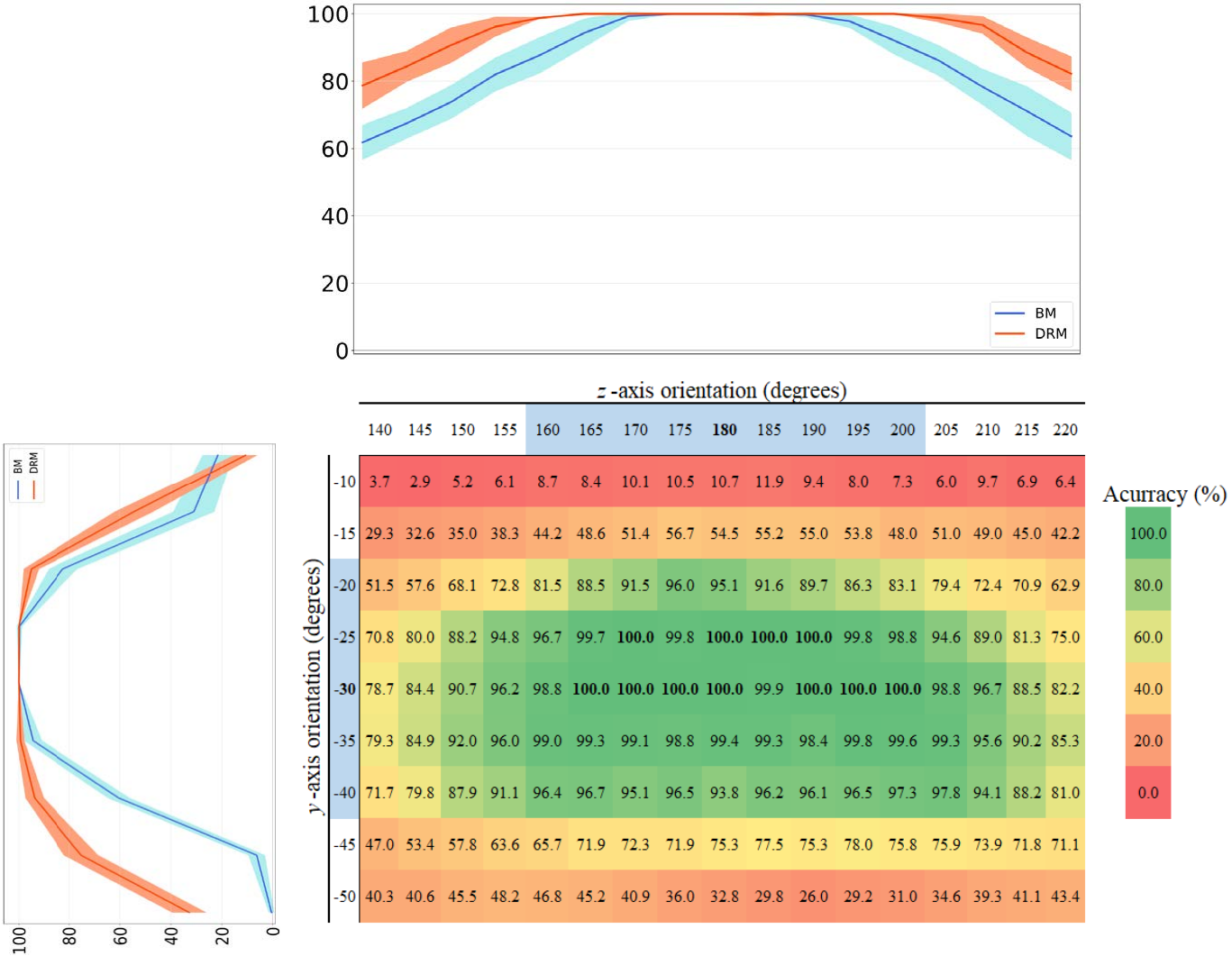}
	\caption{DRM average accuracy obtained in evaluation presented as a heat map. The percentage shown for each camera pose combination is the average of 1,000 test episodes. In its axes, the orientations shown to the DR agent are marked in blue and in bold the single orientation used in the BM. \linebreak On the top, the average accuracy achieved in evaluation across the $z$-axis is presented. It represents the projection of its heat map values when this one is intersected by a perpendicular plane to the $y$-axis at $-30^\circ$. \linebreak On the left, the average accuracy drawn in evaluation across the $y$-axis is showed. It represents the projection of its heat map values when this one is intersected by a perpendicular plane to the $z$-axis at $180^\circ$. \linebreak The line's shading corresponds to $\pm1$ standard deviation of the accuracy distribution per orientation pair.}\label{fig10}
\end{figure}

As in the analysis of the BM, the results of the DRM performance have been analyzed using a representation inspired in the orthographic projection system. The matrix on the center of Fig.~\ref{fig10} displays a heat map whose values are the average accuracy obtained after evaluating the DRM across all the orientation combinations. On top of it, the DRM and BM average accuracy are exhibited along the $z$-axis when the $y$-axis orientation is kept to $-30^\circ$, being this graphics the result of the intersection between each heat map and a plane perpendicular to the $y$-axis at $-30^\circ$. Fig.~\ref{fig10}'s left plot presents the average accuracy when the $z$-axis is fixed at $180^\circ$. This surface projection emulates the intersection between each heat map and a plane perpendicular to the $z$-axis at $180^\circ$.

Owing to the spread experience and thus, the diversity achieved during the learning period, the agent maintain the accuracy above 90\% for approximately all the orientations between $155^\circ$ to $210^\circ$ and $-40^\circ$ to $-25^\circ$, which is clearly a larger range than that in the BM. 100\% accuracy is obtained in 11 positions, most of them at $-30^\circ$ and within the interval $[165^\circ, 200^\circ]$. The maximum failure distance was \SI{81}{cm} at $(140^\circ, -10^\circ)$, one of the most challenging environment configurations for the DRM.

The distribution of the average accuracy around the $z$-axis is more symmetrical than in the BM, especially in the $y$-axis intermediate degrees. Nevertheless, the shadow effect that appeared in the BM causing a considerable decrease in the performance for values near the interval limits did not take place likewise in the DRM. In spite of the 20\% and 15\% accuracy drop at $-45^\circ$ and $-15^\circ$, respectively, the agent learned to manage the shade effect better, leading to a worse performance when the camera orientation is between $[140^\circ, 150^\circ]$ than at $[210^\circ, 220^\circ]$. The agent's operation is definitely better in the DRM than in the BM; nonetheless, it is interesting to notice that the behavior in both models was quite similar since the slopes on the projections are essentially parallel.

When the camera position changed around the $y$-axis, the DRM performance at $-10^\circ$ decreases compared to the BM, and the average accuracy obtained in the worst scenarios is contrary to the previous results. Analyzing the first difference, in Fig.~\ref{fig10} the accuracy is higher at $-50^\circ$ than at $-10^\circ$ for any $z$-degree. We argue that this could be due to the diversity introduced in the model. Since the number of training steps is kept constant, the experience samples are allocated on a broader interval in the DRM than in the BM. Therefore, the Baseline agent is exposed to just one environment configuration that allows it to cope slightly better with the scarcity of depth information, whereas the DR agent, which has seen more different situations but less frequently, presents an inferior performance.

Examining the second divergence, the DR agent presents better accuracy for the orientations around $-50^\circ$ than those at $-10^\circ$. This might be ascribed to the fact that, when variability is introduced during training, it is simpler to learn how to manage situations where the missing information is related to the joints' orientation than to understanding the image depth.

\begin{figure}[t]
	\centering
	\renewcommand{\figurename}{Fig.}
	\includegraphics[width=0.7\textwidth]{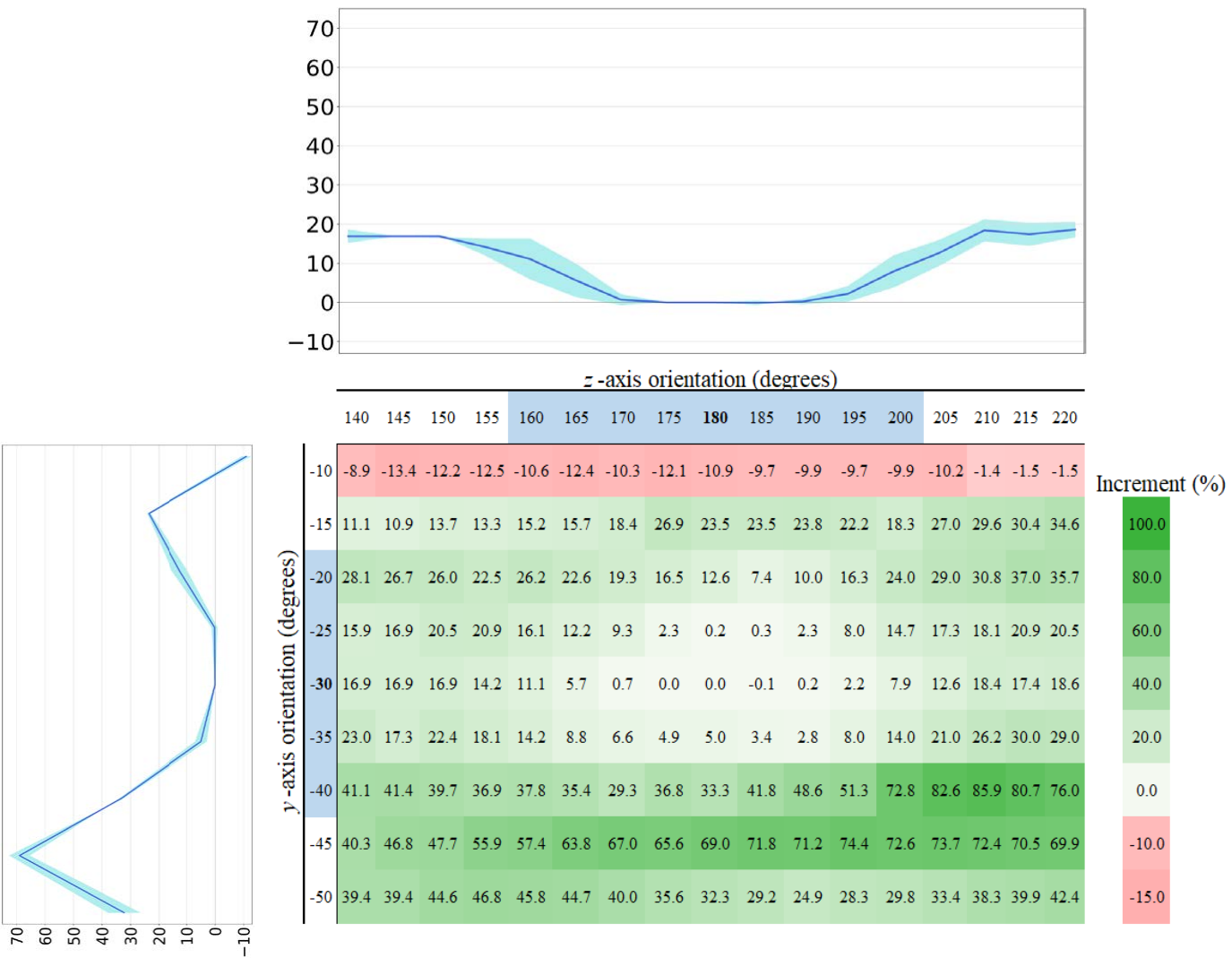}
	\caption{Increment of the average accuracy obtained in evaluation when the DRM is compared with the BM. The percentage shown for each camera pose combination is the average of 1,000 test episodes. In its axes, the orientations shown to the DR agent are marked in blue and in bold the single orientation used in the BM . \linebreak On the top, the increment of the average accuracy achieved in evaluation across the $z$-axis is displayed. It represents the projection of the heat map values when this one is intersected by a perpendicular plane to the $y$-axis at $-30^\circ$. \linebreak On the left, the increment of the average accuracy achieved in evaluation across the $y$-axis is exhibited. It represents the projection of the heat map values when this one is intersected by a perpendicular plane to the $z$-axis at $180^\circ$. \linebreak The line's shading corresponds to $\pm1$ standard deviation of the accuracy distribution per orientation pair.}\label{fig11}
\end{figure}

To conclude, Fig.~\ref{fig11} compares both models by presenting the incremental results obtained for each position, that is, the percentage calculated subtracting the accuracy of the BM from that of the DRM. As in the previous analysis of the BM and the DRM, Fig.~\ref{fig11} is inspired by the orthographic projection system. The top and left plots correspond to the projection of the intersection between the heat maps and a plane perpendicular to the $y$-axis at $-30^\circ$ and the $z$-axis at $180^\circ$ respectively. In contrast, the central matrix represented as a heat map shows all the increment outcomes per camera-pose pair. The surface shows that for orientations near $(180^\circ, -30^\circ)$ the increase is between 0\% and 5\% since the BM already achieved good results in that positions. However, evaluating the rest of the increments reached, we might say that the DRM enhanced the accuracy outcomes significantly, reaching an increment above 80\% in some scenarios and improving accuracy around 25\% on average. The increase for the $z$-axis is distributed as a concave function that reaches 0\% for values in which both models achieved a 100\% accuracy. Nevertheless, the $y$-axis projection has an irregular shape due to the decrease in the DRM's accuracy at $-10^\circ$ and to the outstanding results achieved at $-40^\circ$, where the BM was apparently affected by the robot's shadow. Examining the overall incremental results, it can be asserted that the DRM significantly improves the BM performance within the same training conditions, that is, the same effort, despite being exposed to more variability and, hence, a more complex problem than the BM.

\section{Conclusions and future work}\label{sec7}
\vspace{10pt}
This research compares the performance of an artificial agent trained with domain randomization (DR) techniques with respect to a baseline designed following a vanilla PNN approach. The setup used was a virtual environment composed of a robotic arm whose task is to arrive at a picking area using only the images provided by an externally-mounted camera as input. Both models were exposed to a robustness analysis where the camera orientation changed along two axes. 

The results show that right before being transferred to reality in the sim-to-real pipeline (i.e., after the virtual training phase), the DR agent is more robust to moderate changes in the setup than the Baseline PNN model, at the expense of a slightly higher computational effort, if any. The conclusion holds even if two camera angles are changed simultaneously. This finding suggests that applying DR in the virtual environment should boost the vanilla PNN approach's performance and reduce the amount of training and experience required when transferring this knowledge to the real world, which will never exactly replicate the virtual setup. We leave for future work the quantification of the actual reduction of the total amount of real experience required to make agents fully operative and the point where adding more randomization starts yielding only marginal benefits.

\section*{Author contributions}
\vspace{10pt}
\textbf{Luc\'{i}a G\"{u}itta-L\'{o}pez}: Conceptualization and design of this study, Methodology, Formal analysis and investigation, Software, Data curation, Writing - original draft preparation, Writing - review and editing. \linebreak \textbf{Jaime Boal}: Conceptualization and design of this study, Methodology, Formal analysis and investigation, Supervision, Writing - review and editing.\linebreak \textbf{\'{A}lvaro J. L\'{o}pez-L\'{o}pez}: Conceptualization and design of this study, Methodology, Formal analysis and investigation, Supervision, Writing - review and editing.

\section*{Declarations}
\vspace{10pt}
\begin{itemize}
	\item Funding: This research did not receive any specific grant from funding agencies in the public, commercial, or not-for-profit sectors.
	\item Conflict of interest/Competing interests: All authors certify that they have no affiliations with or involvement in any organization or entity with any financial interest or non-financial interest in the subject matter or materials discussed in this manuscript.
	\item Ethics approval: Not applicable
	\item Consent to participate: Not applicable
	\item Consent for publication: Not applicable
	\item Availability of data and materials: Researchers or interested parties are welcome to contact the corresponding author L.G-L. for further explanation, who may also provide the data upon request.
	\item Code availability: Researchers or interested parties are welcome to contact the corresponding author L.G-L. for further explanation, who may also provide the Python code upon request.
\end{itemize} 

\begin{appendices}
\vspace{10pt}
\section{Design of the Baseline model (BM)}\label{secA1}
\begin{figure}[h!]
	\centering
	\renewcommand{\figurename}{Fig.}
	\includegraphics[width=0.7\textwidth]{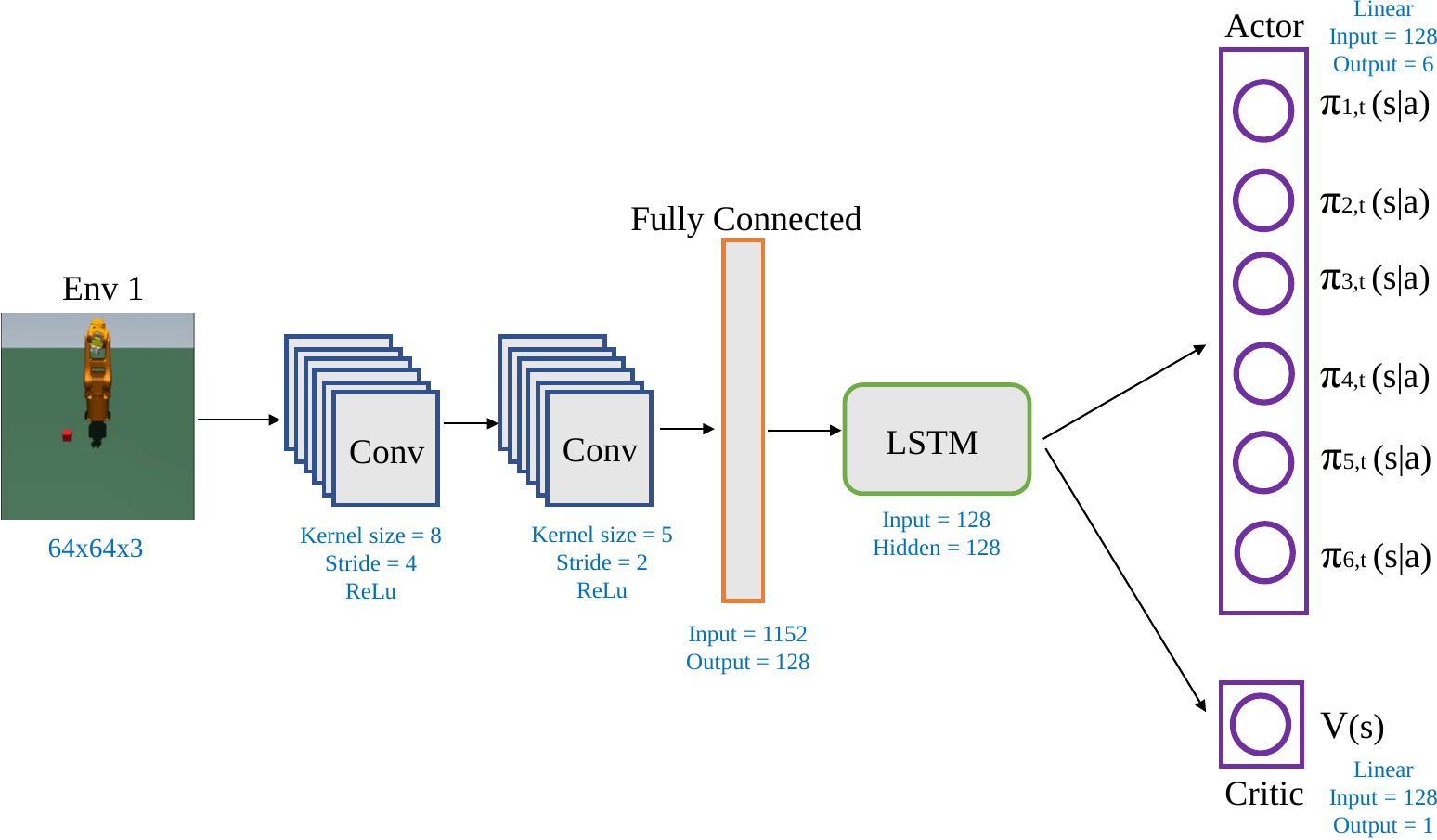}
	\caption{Implemented A3C architecture for the agent. The input is a $64x64$ RGB environment image. Firstly, convolutional layers are applied; the first is defined by a $3x3$ kernel size and $stride = 4$, and the second is characterized by a $5x5$ kernel size and $stride = 2$. The model block is a fully connected layer with 1152 inputs and 128 outputs. Finally, a Long-Short Term Memory (LSTM) network with 128 hidden states is applied to capture better the movements' sequence.}\label{figA1}
\end{figure}

\vspace{10pt}
The problem is solved as a fully observable MDP, where the state signal is the image resulting from rendering the virtual environment. The agent implemented is an A3C whose architecture is the same as that presented by Rusu et al. \cite{Rusu2017}. The agent's observation (i.e., the model input) is a $64x64$ RGB environment image. There are seven outputs, being the Actor composed of six of them, which are the policies applied to each joint, that is, the discrete probability functions that map the probability of applying a certain action from the action-set to modify the joint's position. The softmax function is used to obtain the actions' likelihood. The other model output is the Critic, which determine the value of the value-state function. The network architecture is shown in Fig.~\ref{figA1}.

From the premise that the MDP design was fundamental, various models were developed and evaluated to select the most appropriate action set and reward strategy. In this research, the agent command the position of the actuators rather than their speed, as \cite{Rusu2017} implemented, since that is the attribute embedded controllers allow the user to interact within these types of commercial industrial robots. Hence, the actions are applied as increments of each joints' orientation. Therefore, the proposal made in \cite{Rusu2017} is used as the initial framework due to its encouraging results, but our problem characterization and goals differed. 

In order to analyze the impact of the action sets and the reward on the agent's behavior, several models with different reward and action set configurations were designed and trained. Table~\ref{tabA1}, where $MPI$ stands for Maximum Position Increment and $dist$ for the relative distance between the robot's end-effector and the target and $epLength$ for episode length, shows all the models designed and evaluated. The evaluation of each model is carried out running 1,000 episodes under ten different seeds using first as reward distance \SI{5}{cm} and then \SI{10}{cm}. The metrics analyzed to determine the best model are the average accuracy, the maximum distance failure, and the learning time needed during training.

The Baseline model selected for the remaining research was $M1$. It is characterized by a logarithmic action set that allowed the agent to approach the target faster when it is far away and yet operate precisely in the surroundings of the goal. Regarding the reward, a discontinuous function establishes a negative reinforcement if the goal is not reached and a positive reinforcement otherwise. This configuration is chosen since the outcomes were on a par with the results presented by \cite{Rusu2017}. 
\begin{table}[p]
	\renewcommand{\thefootnote}{\alph{footnote}}
	\begin{center}
		\begin{minipage}{\textwidth}
			\caption{Different MDPs designed and evaluated. \linebreak The Baseline Model selected according to its results in the evaluation is M1.}
\label{tabA1}
			\renewcommand{\arraystretch}{1.5}
			\resizebox{\textwidth}{!}{
			\begin{tabular}{C{2.5cm}C{3.0cm}C{2.0cm}C{2.0cm}}
				Model version & Action space\footnote{$MPI$ stands for Maximum Position Increment.} & Positive reward per step\footnote{$epLength$ stands for episode length.} & Negative reward per step\footnote{$dist$ stands for the relative distance between the gripper and the target.} \\
				\hline
				\hline
				$M0$ & $0$ \linebreak $\pm MPI$ \linebreak $\pm MPI/2$ & $70$ & $-(2*dist)^2$ \\
				\hline
				$M1$ \linebreak \textbf{(Baseline Model chosen)} & $0$ \linebreak $\pm MPI$ \linebreak $\pm MPI/10$ \linebreak $\pm MPI/100$ & $70$ & $-(2*dist)^2$ \\
				\hline
				$M2$ & $0$ \linebreak $\pm MPI$ \linebreak $\pm MPI/10$ \linebreak $\pm MPI/100$ & $90$ if epLength $ \in [0, 30]$ \linebreak $70$ if epLength $\in (30, 35]$ \linebreak $50$ if epLength $\in (35, 40]$ \linebreak $30$ if epLength $\in (40, 50]$ & $-(2*dist)^2$ \\
				\hline
				$M3$ & $0$ \linebreak $\pm MPI$ \linebreak $\pm MPI/10$ \linebreak $\pm MPI/100$ & $100$ if epLength $\in [0, 30]$ \linebreak $50$ if epLength $\in (30, 35]$ \linebreak $30$ if epLength $\in (35, 40]$ \linebreak $20$ if epLength $\in (40, 50]$ & $-(2*dist)^2$ \\
				\hline
				$M4$ & $0$ \linebreak $\pm MPI$ \linebreak $\pm MPI/2$ \linebreak $\pm MPI/4$ \linebreak $\pm MPI/16$ \linebreak $\pm MPI/64$ \linebreak $\pm MPI/128$ & $70$ & $-(2*dist)^2$ \\
				\hline
				$M5$ &  if distance $>$ 1.5*rewarding distance \linebreak $(0, \pm MPI, \pm MPI/2, \linebreak \pm MPI/4)$ \linebreak if distance $<$ 1.5*rewarding distance \linebreak $(0, \pm MPI/10, \pm MPI/50, \linebreak \pm MPI/100)$ & $70$ & $-(2*dist)^2$ \\
				\hline
				$M6$ & if distance $>$ 1.3*rewarding distance \linebreak $(0, \pm MPI, \pm MPI/2, \linebreak \pm MPI/4)$ \linebreak if distance $<$ 1.3*rewarding distance \linebreak $(0, \pm MPI/4, \pm MPI/10, \linebreak \pm MPI/25)$ & $90$ if epLength $\in [0, 30]$ \linebreak $70$ if epLength $\in (30, 35]$ \linebreak $50$ if epLength $\in (35, 40]$ \linebreak $30$ if epLength $\in (40, 50]$ & $-(2*dist)^2$ \\
				\hline
			\end{tabular}
			}
		\end{minipage}
	\end{center}
\end{table}

These experiments conclude that even though the action space discretization can be as granular as desired, simpler spaces lead to better results. Besides, another outcome is that a logarithmic action space seems to behave better than a linear approach, which could be explained by the fact that with the same number of choices, the logarithmic approach provides a better combination of coarse and fine movements.\pagebreak

\end{appendices}

\bibliography{manuscript-bibliography}


\end{document}